%% file: main.tex
\documentclass[conference]{IEEEtran}
\IEEEoverridecommandlockouts
\usepackage{cite}
\usepackage{amsmath,amssymb,amsfonts}
\usepackage{algorithmic}
\usepackage{graphicx}
\usepackage{textcomp}
\usepackage{xcolor}
\def\BibTeX{{\rm B\kern-.05em{\sc i\kern-.025em b}\kern-.08em
    T\kern-.1667em\lower.7ex\hbox{E}\kern-.125emX}}

\usepackage{amsthm}
\usepackage{balance}
\newtheorem{theorem}{Theorem}
\newtheorem{lemma}{Lemma}

\usepackage{commath}
\usepackage{bbm}
\usepackage[ruled,vlined,linesnumbered]{algorithm2e}
\usepackage{booktabs}
\usepackage{multirow}
\usepackage[normalem]{ulem}
\useunder{\uline}{\ul}{}
\usepackage{subfig}
\usepackage{float}
\usepackage{hyperref}
\newcommand{\spara}[1]{\smallskip\noindent{\bf{#1}}}
\newcommand\modelabbrv{NodeSig}
\newcommand\modelname{Binary Node Embeddings via Random Walk Diffusion}

\usepackage{fancyhdr}
\usepackage{kantlipsum}
\fancypagestyle{plain}{
\fancyhf{}
\fancyhead[C]{Conference on \LaTeX} 

}
\usepackage{eso-pic}
\begin{document}

\title{\textsc{\modelabbrv}: \modelname
\thanks{Supported in part by ANR (French National Research Agency) under the JCJC project GraphIA (ANR-20-CE23-0009-01).}
}

\author{\IEEEauthorblockN{Abdulkadir \c{C}elikkanat}
\IEEEauthorblockA{\textit{Technical University of Denmark} \\
\textit{DTU Compute}\\
Lyngby, Denmark \\
abdcelikkanat@gmail.com}
\and
\IEEEauthorblockN{Fragkiskos D. Malliaros}
\IEEEauthorblockA{\textit{Paris-Saclay University} \\ \textit{CentraleSupélec, Inria}\\
Gif-sur-Yvette, France \\
fragkiskos.malliaros@centralesupelec.fr}
\and
\IEEEauthorblockN{Apostolos N. Papadopoulos}
\IEEEauthorblockA{\textit{Aristotle University of Thessaloniki} \\
\textit{Department of Informatics}\\
Thessaloniki, Greece \\
papadopo@csd.auth.gr}

}

\maketitle

\begin{abstract}
Graph Representation Learning (GRL) has become a key paradigm in network analysis, with a plethora of interdisciplinary applications. As the scale of networks increases, most of the widely used learning-based graph representation models also face computational challenges. While there is a recent effort toward designing algorithms that solely deal with scalability issues, most of them behave poorly in terms of accuracy on downstream tasks. In this paper, we aim to study models that balance the trade-off between efficiency and accuracy. In particular, we propose \textsc{\modelabbrv}, a  scalable model that computes binary node representations. \textsc{\modelabbrv} exploits random walk diffusion probabilities via stable random projections towards efficiently computing embeddings in the Hamming space. Our extensive experimental evaluation on various networks has demonstrated that the proposed model achieves a good balance between accuracy and efficiency compared to well-known baseline models on the node classification and link prediction tasks.
\end{abstract}


\input{1-introduction}

\input{2-related_works}
\input{3-method}

\input{4-performance}

\input{5-Discussion}
\input{6-Conclusion}

\bibliographystyle{IEEEtran}
\bibliography{bibliography}

\end{document}

%% file: 1-introduction.tex
\section{Introduction}
\label{sec.intro}

Graph-structured data is ubiquitous in many diverse disciplines and application domains, including biology, neuroscience, and applications arising from social media and networking analysis~\cite{Newman2003}. Besides being elegant models for data representation, graphs have also been proven valuable in various widely used machine learning tasks. For instance, in the case of biological networks, we are interested in predicting the function of proteins or in inferring the missing structure of the underlying protein-protein interaction network. Both of these problems can be formalized as learning tasks on graphs, with the main challenge being how to properly incorporate its structural properties and the proximity among nodes into the learning process. In this direction, \textit{Graph Representation Learning (GRL)} has become a key paradigm for extracting information from networks and for performing various tasks such as link prediction, classification, and visualization  \cite{hamilton-survey17, hamilton-GRL-2020}. These models aim to find node representations (i.e., \textit{node embeddings}) in a way that the desired properties and proximity among nodes are preserved in the embedding space. 

\par Most of the existing GRL approaches deal with \textit{learning-based models}, relying either on matrix factorization or on node context sampling to infer the proximities between nodes \cite{hamilton-survey17}. For the former, the goal is to learn embeddings by \textit{factorizing} the matrix which has been designed for capturing and representing desired graph properties and node proximities in a lower-dimensional space. Typically, such approaches target to preserve first-order (adjacency-based) or higher-order proximity of nodes \cite{cao2015grarep, SDNE-kdd16, mkernelne, HOPE-kdd16}.
Since such models heavily rely on the expensive factorization of dense node proximity matrices, the computational cost and the high memory usage burden bring limitations 
for large-scale networks. Although recent studies have proposed heuristics to overcome these challenges \cite{ZCLW+18, CSTC+19}, they are obliged to forgo their predictive performance in most cases---hence, putting the practitioners in a dilemma between effectiveness and 
computational cost. 


\par In order to address the aforementioned challenges towards developing effective and scalable algorithms for representation learning on networks, \textit{random walk}-based models have gained considerable attention \cite{hamilton-survey17}. The main idea here is to generate a set of node sequences by following a random walk strategy. Node embeddings are then learned by maximizing the probability of node co-occurrences in the generated sequences. \cite{PAS14, GL16, TQWZ+15, biasedwalk, efge-aaai20}. 
Nevertheless, 
a large number of random walks is required to be explicitly sampled in order to ensure the effectiveness of the embedding on downstream tasks. 
Furthermore, it has been shown that random walk-based embedding approaches 
implicitly perform factorization of a properly chosen dense transition probability matrix, leading to better performance on downstream tasks \cite{QDML+18, infinitewalk-kdd20}. Although recent studies aim to improve running time complexity via matrix sparsification techniques \cite{netsmf-www19} or capitalizing on hierarchical graph representations \cite{louvainNE-wsdm20},  the quality of the embeddings deteriorates.

\par Besides the computational burden of model optimization, most of the proposed algorithms learn low-dimensional embeddings in the \textit{Euclidean} space. 
A recent few studies have proposed to learn \textit{discrete} node  representations \cite{inh-mf-kdd18, WLCZ18}, in which \textit{Hamming} distance is leveraged to determine the similarity of embedding vectors. The basic idea builds upon fast sketching techniques for scalable similarity search, mainly based on data-independent or data-dependent hashing techniques \cite{learning-to-hash}. Although binary embeddings speedup distance measure computations with respect to the metrics defined in Euclidean space, the corresponding models often undergo computationally intensive learning procedures, especially in the case of learning-to-hash models \cite{inh-mf-kdd18}. 

\spara{Contributions}. In this paper, we propose \textsc{\modelabbrv}, a scalable model for computing expressive binary node embeddings based on stable random projections.  \textsc{\modelabbrv} first leverages random walk diffusions to estimate higher-order node proximity. Then, a properly defined sign random projection hashing technique is applied to obtain binary node signatures in the Hamming space, leading to an approximation of the \textit{chi similarity} ($\chi$) \cite{pele_wermen} between the proximity vectors in the original space. Since these vectors are constructed based on the occurrence frequencies of nodes within random walks, \textit{chi} similarity emerges as a natural choice of similarity metric, frequently used to compare histograms in various areas including natural language processing and computer vision \cite{chen_jian_victor_zhao, huong_park_woo}. 

Each component of \textsc{\modelabbrv} has been designed to ensure the scalability, while at the same time, the accuracy on downstream tasks is not compromised or even improves compared to traditional models.  Figure~\ref{fig:method_comparison} positions \textsc{\modelabbrv} regarding the accuracy and running time, providing a comparison to different models on the \textsl{PPI} network. As we observe, \textsc{\modelabbrv}'s running time is comparable to that of models that focus solely on scalability (e.g., \textsc{NodeSketch, RandNE, LouvainNE}), with improved accuracy even higher than \textsc{Node2Vec}, \textsc{FREDE} and \textsc{HOPE} in this dataset. 

\noindent The main contributions can be summarized as follows:
\begin{itemize}
    \item We introduce \textsc{\modelabbrv}, a scalable and expressive model for binary node embeddings based on stable random projection hashing of random walk diffusion probabilities.
    
    \item The distance computation between node signatures in the embedding space is provided by the Hamming distance on bit vectors, which is significantly more efficient than distance computations based on other distance measures.
    
    \item In a thorough experimental evaluation, we demonstrate that the proposed binary embeddings achieve superior performance compared to various baseline models on two downstream tasks. At the same time, the running time 
    allows the model to scale on large graphs.
\end{itemize}

\spara{Source code}. 
The implementation of \textsc{NodeSig} can be found at: \url{https://abdcelikkanat.github.io/projects/nodesig/}.

%% file: 2-related_works.tex
\section{Related Work}
\label{sec.related}

\begin{figure}
\centering
\includegraphics[width=0.7\columnwidth]{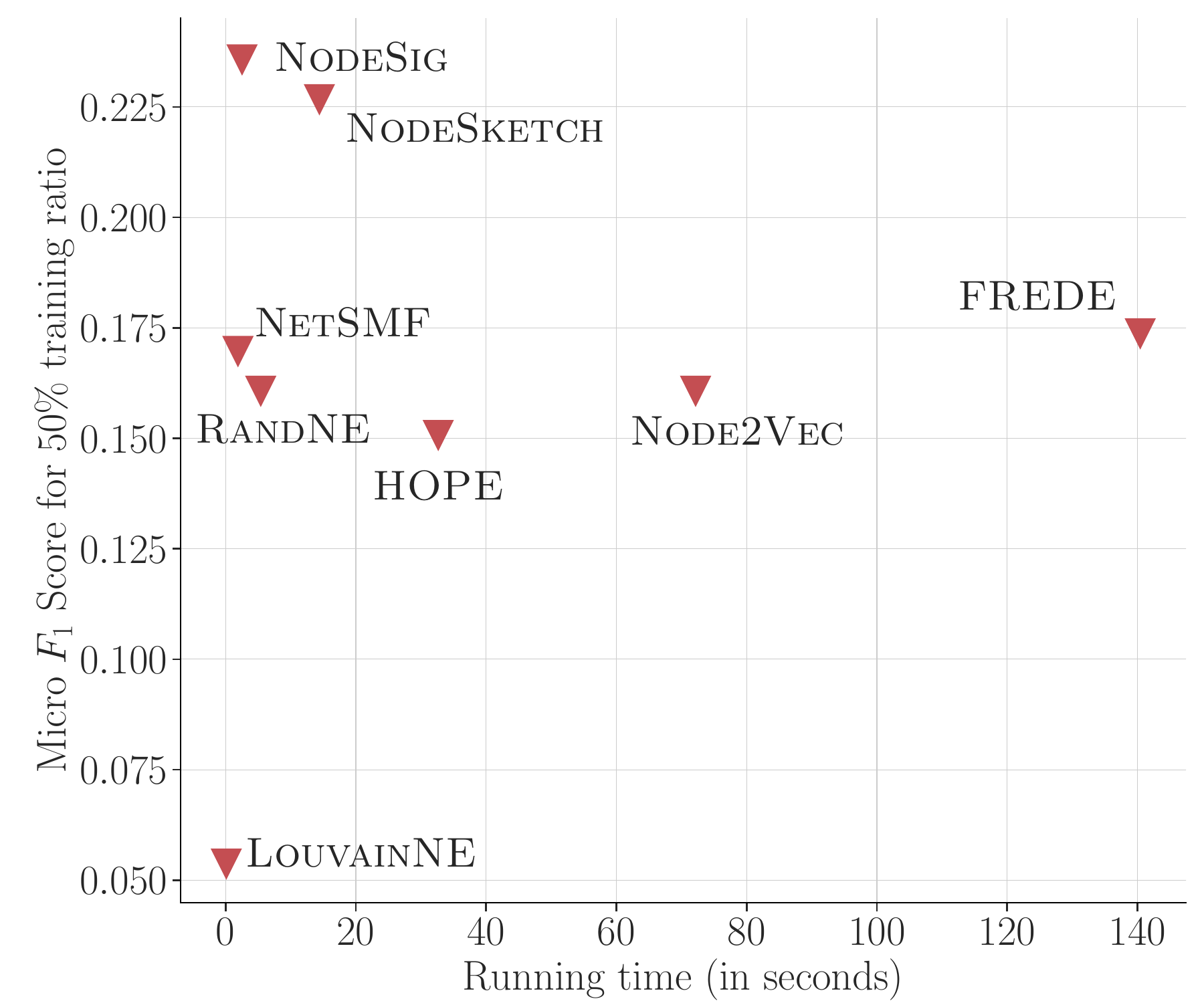}
\caption{Comparison of models on the \textsl{PPI} network. \textsc{\modelabbrv} balances good accuracy (Micro-$F_1$) and running time (in secs). \vspace{-.3cm}}
\label{fig:method_comparison}
\end{figure}

Graphs are rich representations of the real world that can capture different types of relationships and modalities among entities \cite{hamilton-survey17}. 
One of the first modern algorithms is the \textsc{DeepWalk} algorithm \cite{PAS14} which uses uniform truncated random walks to represent the \textit{context} of a node. Intuitively, nodes with similar random walks have a higher degree of similarity. 
The \textsc{Node2Vec} method \cite{GL16} is more general and manages to combine BFS and DFS search strategies, achieving significant performance improvements. 
\textsc{LINE} \cite{TQWZ+15} optimizes an objective function capturing both first-order and second-order node proximities.
Essentially those models constitute adaptations of the \textsc{SkipGram} technique proposed for word embeddings \cite{MSCC+13}.
\textcolor{black}{However, they require the extensive realization of the random walks, which constitutes a computationally intensive operation.} It turns out that problems related to random walk sampling can be alleviated by using matrix factorization. 
The main drawback of \textsc{NetMF} and other matrix factorization-based approaches \cite{cao2015grarep, HOPE-kdd16}, however, is that in general matrix factorization is a computationally intensive operation.

The main limitation of the aforementioned embedding techniques is that they do not scale well for large networks. The main focus has been put on increasing the effectiveness of data mining tasks (e.g., classification, link prediction, network reconstruction) whereas the efficiency dimension has not received significant attention. To attack this problem, recent advances in network representation learning use random projection or hashing techniques (more specifically, variants of locality-sensitive hashing) in order to boost performance, trying to maintain effectiveness as well.  

\textsc{RandNE} \cite{ZCLW+18} is one of the first scalable approaches which is based on iterative Gaussian random projection, being able to adapt to any desired proximity level. 
In the same line, \textsc{FastRP} was proposed in~\cite{CSTC+19} which is faster than \textsc{RandNE} and also more accurate. \textsc{LouvainNE} \cite{louvainNE-wsdm20} suggested learning node representations by aggregating the embeddings of nodes extracted at varying levels of the hierarchy. 
The \textsc{NetHash} algorithm \cite{WLCZ18} expands each node of the graph into a rooted tree, and then by using a bottom-up approach encodes structural information as well as attribute values into minhash signatures in a recursive manner. \textsc{FREDE} \cite{frede} is a sketching-based approach relying on Personalized Page Rank (PPR) matrix, which alleviates the computation burden by applying a sketching technique.
A similar approach has been used in \textsc{NodeSketch}~\cite{YRLC19}, where the context of every node is defined in a different way whereas the embedding vector of each node contains integer values. \textcolor{black}{Although these approaches rely on fast-sketching schemes, they do not show comparable performance to the aforementioned learnable models in the downstream tasks. In this paper, we aim to introduce an approach balancing accuracy and running time.}


%% file: 3-method.tex
\begin{figure*}[t]
\centering
\includegraphics[width=0.885\textwidth]{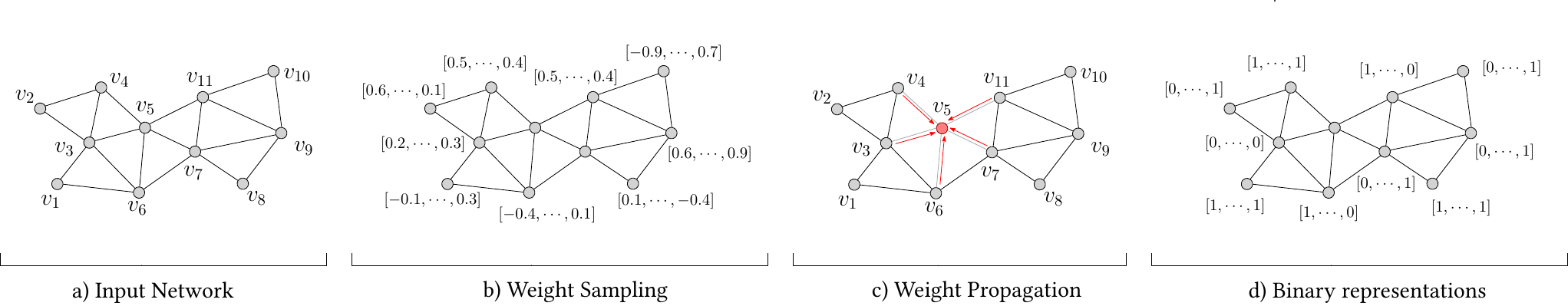}
\caption{Schematic representation of the \textsc{\modelabbrv} model. First, the weights of the random projection matrix are sampled and then the projection of the proximity matrix is performed via the weight propagation step. Finally, binary node representations are obtained by combining the signs of the projected values.}
\label{fig:overview}
\end{figure*}
\section{Proposed Approach}
\label{sec.proposed}
\subsection{Random Walk Diffusion for Node Proximity Estimation}
In most cases, direct links among nodes are not sufficient to grasp various inherent properties of the network that are related to node proximity. It is highly probable that the network might have missing or noisy connections, thus relying solely on first-order proximity can reduce the expressiveness of the model. To overcome this problem, we directly leverage random walk diffusions, adopting a uniform random walking strategy to extract information describing the structural roles of nodes in the network. Let $\mathbf{P}$ denote the right stochastic matrix associated with the adjacency matrix of the graph, which is obtained by normalizing the rows of the matrix. More formally, $\mathbf{P}$ can be written as $\mathbf{P}_{(i,j)} := \mathbf{A}_{(i,j)}/\sum_{j}\mathbf{A}_{(i,j)}$, defining the transition probabilities of the uniform random walk strategy. We use a slightly modified version of the transition matrix by adding a self-loop on each node, in case it does not exist. 

Note that the probability of visiting the next node depends only on the current node that the random walk resides; therefore, node $v_j$ can be visited starting from $v_i$ by taking $l$ steps with probability $\mathbf{P}^{(l)}_{(i,j)}$, if there is a path connecting them. For a given walk length $L$,  we define the matrix $\mathbf{M}$ as
\begin{align*}
    \mathbf{M} := \mathbf{P} +  \cdots + \mathbf{P}^{(l)} + \cdots + \mathbf{P}^{(L)},
\end{align*}
\noindent
where $\mathbf{P}^{(l)}$ indicates the $l$-order proximity matrix and each entry $\mathbf{M}_{v,u}$ in fact specifies the expectation of visiting $u$ starting from node $v$  within $L$ steps. By introducing an additional parameter $\alpha$, $\mathbf{M}(\alpha)$ can be rewritten as follows:
\begin{align*}
    \mathbf{M}(\alpha) := \alpha\mathbf{P} +  \cdots + \alpha^{(l)}\mathbf{P}^{(l)} + \cdots + \alpha^{(L)}\mathbf{P}^{(L)}.
\end{align*}
\noindent Higher order node proximities can be captured  using longer walk lengths, where the impact of the walk at different steps is controlled by the \textit{importance factor} $\alpha\in\mathbb{R}^+$. As we will present in the next paragraph, matrix $\mathbf{M}(\alpha)$ is properly exploited by a random projection hashing strategy to efficiently compute binary node representations.

\subsection{Learning Binary Embeddings}
Random projection methods \cite{vempala2001random} have been widely  used in a wide range of machine learning applications dealing with large scale data. They mainly target to represent data points into a lower dimensional space by preserving the similarity in the original space. Likewise, we aim at encoding each node into a \textit{Hamming} space $\mathbb{H}\big(d_{\mathcal{H}}, \{0,1\}^{\mathcal{D}}\big)$; we consider the normalized \textit{Hamming} distance $d_{\mathcal{H}}$ as the distance metric \cite{binary_emb}. The benefit of binary representations is twofold: first, they will allow us to perform efficient distance computation using bitwise operations, and secondly reduce the required disk space to store the data.

Random projections are linear mappings; the binary embeddings though require  nonlinear functions to perform the discretization step, and a natural choice is to consider the signs of the values obtained by the \textit{Johnson-Lindenstrauss (JL)} \cite{Johnson1986} transform. More formally, it can be written that
\begin{equation*}
h_{\mathbf{W}}(\mathbf{x}) := sign(\mathbf{x}^{\top}\mathbf{W}),    
\end{equation*}
\noindent where $\mathbf{W}$ is the projection matrix whose entries $\mathbf{W}_{(i,j)}$ are independently drawn from normal distribution and  $sign(\mathbf{x})_j$ is equal to $1$ if $\mathbf{x}_j>0$ and $0$ otherwise.  The approach was first introduced in the work \cite{GW95} for a rounding scheme in approximation algorithms, demonstrating that the probability of obtaining different values for a single bit quantization is proportional to the angle between vectors, as it is shown in Theorem \ref{thm:projection}. The main idea relies on sampling uniformly distributed random hyperplanes in $\mathbb{R}^{\mathcal{D}}$. Each column of the projection matrix, in fact, defines a hyperplane 
and the arc between vectors $\mathbf{x}$ and $\mathbf{y}$ on the unit sphere is intersected if $h_{\mathbf{W}}(\mathbf{x})_i$ and  $h_{\mathbf{W}}(\mathbf{y})_i$ take different values. 

\begin{theorem}[\cite{GW95}]\label{thm:projection}
For a given pair of vectors $\mathbf{x}, \mathbf{y} \in \mathbb{R}^{N}$,
\begin{align*}
    \mathbb{P}\left[h_{\mathbf{W}}(\mathbf{x})_j \not= h_{\mathbf{W}}(\mathbf{y})_j \right] = \frac{1}{\pi}\text{\normalfont cos}^{-1}\left(\frac{\mathbf{x}^{\top} \mathbf{y}}{\Vert\mathbf{x}\Vert_2\Vert\mathbf{y}\Vert_2}\right),
\end{align*}
\noindent where $\mathbf{W}_{(i,j)} \sim \mathcal{N}(0,1)$ for $1\leq i,j \leq N$.
\end{theorem}

Although the signs of JL random projections allow us to approximate the angle between the vectors in the original space, in our settings, we would prefer to preserve a distance metric that can fit better the input data $\mathbf{M}(\alpha)$. Note that, the node proximity matrix $\mathbf{M}(\alpha)$ contains non-negative elements computed based on the occurrence frequencies of nodes within random walks. Hence, we will focus on estimating distance metrics capable of comparing histogram-type data by properly redesigning the projection matrix. The \textit{stable} random projections approach \cite{sign_cauchy_proj} generalizes the aforementioned idea by using a symmetric $\alpha$-stable distribution with unit scale in order to sample the elements of the projection matrix, for $0 < \alpha \leq 2$. Li et al. \cite{sign_cauchy_proj} proposed the following upper bound
\begin{align}
    \mathbb{P}\left[h_{\mathbf{W}}(\mathbf{x})_j \not= h_{\mathbf{W}}(\mathbf{y})_j \right] \leq \frac{1}{\pi}\text{\normalfont cos}^{-1}\rho_{\alpha}
\end{align}
for non-negative vectors ($\mathbf{x}_i \geq 0, \mathbf{y}_i \geq 0$ for $1 \leq i \leq N$), where $\rho_{\alpha}$ is defined as
\begin{align*}
\rho_{\alpha} := \left( \frac{ \sum_{i=1}{x}_i^{\alpha/2}{y}_i^{\alpha/2} }{ \sqrt{\sum_{i=1}{x}_i^{\alpha} }\sqrt{\sum_{i=1}{y}_i^{\alpha} } } \right)^{2/\alpha}.
\end{align*}

\noindent It is well known that the bound is exact for $\alpha = 2$, which also corresponds to the special case in which normal random projections are performed. When the vectors are chosen from the $\ell^1(\mathbb{R}^+)$ space (i.e.,  $\sum_{d=1}x_d=1$, $\sum_{d=1}y_d=1$), it is easy to see that the $\chi^2$ similarity  $\rho_{\chi^2}$ defined as $\sum_{d=1}(2x_dy_d)/(x_d + y_d)$ is always greater or equal to $\rho_1$, as suggested by Lemma \ref{lem}. 

\begin{lemma}\label{lem}
For given $\mathbf{x}, \mathbf{y} \in \mathbb{R}^{\mathcal{D}}$ satisfying $x_i, y_i\geq 0$ for all $1\leq i \leq \mathcal{D}$ and $\sum_{i=1}x_i=\sum_{i=1}y_i=1$, then $\rho_1 \leq \rho_{\chi^2}$.
\end{lemma}
\begin{proof}
\begin{align*}
\rho_1 \!\!=\! \left(\frac{ \sum_{i=1}{x}_i^{1/2}{y}_i^{1/2} }{ \sqrt{\sum_{i=1}{x}_i }\sqrt{\sum_{i=1}{y}_i } }\right)^2  \!\!\!\!&=\! \left( \sum_{i=1}\sqrt{{x}_i{y}_i}\right)^2 \!\!\!\!
\\
&=\! \left(\sum_{i=1}\frac{\sqrt{2x_iy_i}}{\sqrt{x_i+y_i}}\frac{\sqrt{x_i+y_i}}{\sqrt{2}}\right)^2 \\
&\leq \sum_{i=1}\frac{2x_iy_i}{x_i+y_i} \sum_{i=1}\frac{x_i+y_i}{2} \\
&=\sum_{i=1}\frac{2x_iy_i}{x_i+y_i} = \rho_{\chi^2},
\end{align*}
where the inequality follows from the \textit{Cauchy-Schwarz} inequality.
\end{proof}
Besides, it has been empirically shown \cite{sign_cauchy_proj} that the collision probability for \textit{Cauchy} random projections with unit scale can be well estimated, especially for sparse data:
\begin{align}\label{eq:main}
    \mathbb{P}\left[h_{\mathbf{W}}(\mathbf{x})_j \not= h_{\mathbf{W}}(\mathbf{y})_j \right] \approx \frac{1}{\pi}\text{\normalfont cos}^{-1}\rho_{\chi^2} \leq \frac{1}{\pi}\text{\normalfont cos}^{-1}\rho_{1}. 
\end{align}
\noindent Note that, the matrix, $\mathbf{M}(\alpha)$, described in the previous paragraph consists of non-negative values; its row sums are equal to $\sum_{l=1}^L\alpha^l$ and $\mathbf{M}(\alpha)$ is sparse enough for small walk lengths. Therefore, we design the projection matrix by sampling its entries from the \textit{Cauchy} distribution, aiming to learn binary representations preserving the \textit{chi-square} similarity. The \textit{chi-square} distance is one of the measures used for histogram-based data, commonly used in the fields of computer vision and natural language processing \cite{chen_jian_victor_zhao, huong_park_woo}.

As it is shown in Figure~\ref{fig:overview}, the last step of \textsc{\modelabbrv} for obtaining binary node representations is to utilize the signs of the projected data. In other words, the embedding vector $ \mathbf{E}[v]$ for each node $v\in V$ is computed as follows:
\begin{align*}
\mathbf{E}[v]\!:= \!\!\left[sign\!\left(\!\mathbf{M}(\alpha)_{(v,:)} \ \!\!\!\mathbf{W}_{(:,1)}\!\right),\ldots, sign\!\left(\!\mathbf{M}(\alpha)_{(v,:)}\ \!\!\mathbf{W}_{(:,\mathcal{D})}\!\right)\!\right]
\end{align*}

\noindent Note that, the projection of the exact realization of $\mathbf{M}(\alpha)$ can be computationally intensive, especially for large walks. Instead, it can be computed by propagating the weights ${W}_{(u, d)}$ for each dimension $d$ ($1\leq d \leq \mathcal{D}$), using the following recursive update rule: 
\begin{align}\label{eq:update_rule}
    \mathcal{R}^{(l+1)}_{(v,d)}(\alpha) \gets  \alpha \sum_{u\in\mathcal{N}(v)} {P}_{(v, u)} \times \left( {W}_{(u, d)} + \mathcal{R}^{(l)}_{(u,d)}(\alpha) \right),
\end{align}

\noindent
where $\mathcal{N}(v)$ refers to the set of neighbors of node $v\in V$ and $\mathcal{R}^{(l)}_{(v,d)}$ is equal to the projected data, $\mathbf{M}_{(v,:)}(\alpha)\cdot\mathbf{W}_{(:,d)}$ for the walk length $l$, and $\mathcal{R}^{(0)}_{(v,d)}$ is initialized to zero. By Lemma \ref{lemma:recursive_matrix_multiplication}, it can be seen that the projection of $\mathbf{M}(\alpha)$ can be computed by applying the recursive update rules defined in Eq. \ref{eq:update_rule}.

\begin{lemma}\label{lemma:recursive_matrix_multiplication}
Let $\mathbf{P}$ be $n\times n$ a right stochastic matrix and $\mathbf{M}^{(L)}(\alpha)$ be the matrix defined by $ \alpha\mathbf{P} +  \cdots + \alpha^{(l)}\mathbf{P}^{(l)} + \cdots + \alpha^{(L)}\mathbf{P}^{(L)}$. 
For a given $\mathbf{W} \in \mathbb{R}^{n\times\mathcal{D}}$, the term $\mathbf{M}^{(L)}(\alpha)\mathbf{W}$ is equal to $\mathcal{R}^{(L)}$
where each $R^{(l)}_{(v,d)}$ is recursively defined by $\alpha\mathbf{P}_{(v,:)}\big(\mathbf{W}_{(:,d)} + \mathcal{R}^{(l-1)}_{(:,d)}\big)$ for all $l\in\{1,\dots,{L}\}$, $v\in\{1,\ldots,n\}$ and $\mathcal{R}^{(0)}$ is set to $0$. 
\end{lemma}
\begin{proof}
For $l=1$, we have that $\mathcal{R}^{(1)}_{(v,d)} = \alpha \mathbf{P}_{(v,:)}\big( \mathbf{W}_{(:,d)} + \mathcal{R}^{(0)}_{(:,d)} \big) = \alpha \mathbf{P}_{(v,:)}\big( \mathbf{W}_{(:,d)} + 0\big) =  \alpha \mathbf{P}_{(v,:)}\mathbf{W}_{(:,d)}$ for all $d\in\{1,\ldots,\mathcal{D}\}$, $v\in\{1,\ldots,n\}$ so the claim holds for $n=1$. Let us assume that it is true for $n=l\geq1$. Then,
	\begin{align*}
		\mathcal{R}^{(n+1)} _{(v,d)}&= \alpha \mathbf{P}_{(v,:)}\!\!\left( \mathbf{W}_{(:,d)} + \mathcal{R}^{(n)}_{(:,d)} \right)\\
		&=  \alpha \mathbf{P}_{(v,:)}\!\!\left(\!\mathbf{W}_{(:,d)}\!+\! \alpha \mathbf{P} \mathbf{W}_{(:,d)} + \cdots + \alpha^{(n)} \mathbf{P}^{(n)} \mathbf{W}_{(:,d)}\!\right)\\
		&= \alpha \mathbf{P}_{(v,:)} \mathbf{W}_{(:,d)} + \cdots + \alpha^{(n+1)} \mathbf{P}^{(n+1)}_{(v,:)} \mathbf{W}_{(:,d)} \\
		&= \left( \alpha \mathbf{P}_{(v,:)} + \cdots + \alpha^{(n+1)} \mathbf{P}^{(n+1)}_{(v,:)} \right) \mathbf{W}_{(:,d)} \\
		&= \mathbf{M}^{(n+1)}(\alpha)\mathbf{W}_{(:,d)}.
	\end{align*}
	\noindent Thus, the claim also holds for $n\!+\!1\!=\!l$. By the principle of induction, it satisfies for all $l \in \{1, \ldots,{L}\}$.
\end{proof}

\noindent Algorithm~\ref{alg:\modelabbrv} provides the pseudocode of \textsc{\modelabbrv}. We generate the projection matrix by sampling the weights from the \textit{Cauchy} distribution with unit scale. The samples are further divided by $\sum_{l=1}\alpha^l$, because the row sums of $\mathbf{M}(\alpha)$ must be equal to $1$. Then, we compute the terms $\mathcal{R}_{(v,d)}^{(l)}$ by propagating the weights in Line \ref{alg:alg_update_rule} at each walk iteration $l<L$. Note that, the term $R$ in the pseudocode is a vector of length $\mathcal{D}$, thus we obtain the final node representation using the signs of $\mathcal{R}_{(v,d)}^{(L)}$. 

\begin{algorithm}[t]
\SetAlgoLined
\KwData{Graph $\mathcal{G}=(V,E)$ with the transition matrix $\mathbf{P}$; representation size $\mathcal{D}$; walk length $L$; importance factor $\alpha$}
\KwResult{Embeddings $\mathbf{E}[v]\in\mathbb{R}^{\mathcal{D}}$ for each node $v\in V$}
\For{\textbf{each} node $v \in V$}{
$\mathcal{R}[v] \gets \mathbf{0}_{\mathcal{D}}=(0,\ldots,0)$ \;
$W[v] \sim Cauchy(0,1)^{\mathcal{D}}$ / $\sum_{l=1}^{L}\alpha^l$\;
}
\For{$l \gets 1$ $\KwTo$ $L$}{
\For{\textbf{each} node $v \in V$}{\label{alg:alg_loop}
$temp[v] \gets \mathbf{0}_{\mathcal{D}}=(0,\ldots,0)$ \;
\For{\textbf{each} neighbour node $u\in \mathcal{N}(v)$}{
$temp[v] \!\gets temp[v]\! +\! \left( W[u]\! +\! \mathcal{R}[u] \right)\! \times\! P[u,v]$\;\label{alg:alg_update_rule}
}
}
\For{\textbf{each} node $v \in V$}{
$ \mathcal{R}[v] \gets \alpha \times temp[v]$ \;
}
}
\For{\textbf{each} node $v \in V$}{
$ \mathbf{E}[v] \gets sign(\mathcal{R}[v])$ \;
}
\caption{\textsc{\modelabbrv}}\label{alg:\modelabbrv}
\end{algorithm}

\subsection{Time and Space Complexity}
At the beginning of the algorithm, we need to sample a weight matrix of size $|\mathcal{V}|\cdot \mathcal{D}$, and it can be formed in the order of $\mathcal{O}(|\mathcal{V}|\cdot\mathcal{D})$. As we observe in Algorithm~\ref{alg:\modelabbrv}, 
the main cumbersome point of \textsc{\modelabbrv} is caused by the update rule defined in Eq. \eqref{eq:update_rule}, which corresponds to Line \ref{alg:alg_update_rule} of the pseudocode. 
The update rule must be repeated $|\mathcal{N}(v)|$ times for each node $v\in V$, thus it requires $2 \cdot m \cdot \mathcal{D}$ multiplication operations at the walk step $l$ ($1 \leq l \leq L$) for a network consisting of $m$ edges and for embedding vectors of dimension $\mathcal{D}$. Hence, the overall time complexity of the algorithm is $\mathcal{O}\left( (|\mathcal{V}|\cdot\mathcal{D} + m \cdot L \cdot \mathcal{D}\right)$. During the running course of the algorithm, we need to store a vector of size $N$ in memory for the computation of each dimension. Assuming, in the worst case, that we aim to retain the whole projection matrix $\mathbf{W}$ in memory, we need $\mathcal{O}(N \cdot \mathcal{D})$ space in total, since each node requires $\mathcal{D}$ space for storing the $\mathcal{R}_{(v,d)}^{(l)}$ values in the update rule of Eq. \eqref{eq:update_rule}. Note that, the performance of the algorithm  can be boosted by using parallel processing for each dimension of embedding vectors or for Line \ref{alg:alg_loop}, since the required computation for each node is completely independent.

%% file: 4-performance.tex
\section{Experimental Evaluation}
\label{sec.performance}
We report empirical evaluation results demonstrating the effectiveness and efficiency of \textsc{\modelabbrv} compared  to baselines. All the experiments have been performed on a server ($16$ Cores) with $128$GB of memory. 

\subsection{Datasets and Baseline Models}\label{subsec:alg_datasets}
\noindent \textbf{Datasets.} We perform  experiments on networks of different scale and type. $(i)$ \textsl{Blogcatalog} \cite{blogcatalog} social network;  $(ii)$ \textsl{Cora} \cite{cora} citation graph; $(iii)$ \textsl{DBLP} \cite{dblp}  co-authorship network; $(iv)$ \textsl{PPI} \cite{GL16} is a protein-protein interaction network. $(v)$ \textsl{Youtube} \cite{youtube} is a social network in which node labels indicate categories of videos. All the networks used in the experiments are unweighted and undirected (the direction of edges are discarded), in order to be consistent in the evaluation. The characteristics of the graphs are shown in Table \ref{tab:network-statistics}.

\noindent \textbf{Baseline models.} We have considered seven representative baseline methods in the evaluation.  In particular,  the first two correspond to widely used node embedding models: $(i)$ a biased random walk-based model, \textsc{Node2Vec} \cite{GL16}, and $(ii)$ a matrix factorization algorithm, \textsc{HOPE} \cite{HOPE-kdd16}. The remaining four baselines constitute recent models aiming to address the scalability challenge. $(iii)$ \textsc{NetSMF} \cite{netsmf-www19} is a sparse matrix factorization algorithm, modeling the pointwise mutual information of node co-occurrences. $(iv)$ \textsc{FREDE} \cite{frede} is a matrix sketching-based approach. $(v)$ \textsc{RandNE} \cite{ZCLW+18} leverages Gaussian random projections to deal with scalability. $(vi)$ \textsc{LouvainNE} \cite{louvainNE-wsdm20} constructs a hierarchical subgraph structure, aggregating the node representations learned at each level. Finally, $(vii)$ \textsc{NodeSketch} \cite{YRLC19} learns embeddings in the \textit{Hamming} space, using \textsc{MinHash} signatures. 
For all methods, we learn embedding vectors of size $128$.

\par For simplicity, we set the importance factor $\alpha$ to $1$ in all the experiments of \textsc{\modelabbrv}, as we have observed that the algorithm shows comparable performance for values close to $1$; a detailed analysis of the behaviour of \textsc{\modelabbrv} with respect to  the importance factor is given in Section \ref{subsection:param_sensitivity}. The walk length is set to $3$ for \textsl{Cora} and \textsl{Blogcatalog}, and to $5$ for all the other networks in the classification experiment. For the link prediction task, the walk length is chosen as $15$ for all networks. We set the dimension size of the embedding vectors to $8,192$ bits in order to be consistent with the experiments with the baseline methods, since modern computer architectures use $8$ Bytes for storing floating point data types. 


\begin{table}[]
\centering
\begin{tabular}{rcccc}
\toprule
\textbf{Networks} & \textbf{\# Nodes} & \textbf{\# Edges} & \textbf{\# Labels} & \textbf{\# Density} \\ \midrule
\textsl{Blogcatalog} & 10,312 & 333,983 & 39 & $6.3\times 10^{-3}$ \\
\textsl{Cora} & 2,708 & 5,278 & 7 & $1.4\times 10^{-3}$ \\
\textsl{DBLP} & 27,199 & 66,832 & 4 & $1.8\times 10^{-4}$ \\
\textsl{PPI} & 3,890 & 38,739 & 50 & $5.1 \times 10^{-3}$ \\
\textsl{Youtube} & 1,138,499 & 2,990,443 & 47 & $4.6\times 10^{-6}$ \\ \bottomrule
\end{tabular}%
\caption{Characteristics of networks.}
\label{tab:network-statistics}
\end{table}

\subsection{Multi-label Node Classification}\label{subsec:classification}
Our goal is to correctly infer the labels of nodes chosen for the testing set, using the learned representations and the labels of nodes in the rest of the network, namely the nodes in the training set. The evaluation follows a strategy similar to the one used by baseline models \cite{YRLC19}. 

\subsubsection{Experimental set-up}
The experiments are carried out by training an one-vs-rest \textit{SVM} classifier with a pre-computed kernel, which is designed by computing the similarities of node embeddings. The similarity measure is chosen depending on the algorithm that we use to learn representations. More specifically, the \textit{Hamming} similarity for \textsc{NodeSketch} and  the \textit{Cosine} similarity for the rest baselines methods are chosen in order to build the kernels for the classifier. For \textsc{\modelabbrv}, we use the \textit{chi} similarity $\chi$, defined as $1 - \sqrt{d_{\chi^2}}$, where 
\begin{align*}
    d_{\chi^2} \!:=\! \sum_{i=1}^{\mathcal{D}}\frac{({x}_i-{y}_i)^2}{{x}_i+{y}_i} \!=\! \sum_{i=1}^{\mathcal{D}}({x}_i+{y}_i) \!-\! \sum_{i=1}^{\mathcal{D}}\frac{4{x}_i{y}_i}{{x}_i+{y}_i} \!=\! 2 - 2\rho_{\chi^2},
\end{align*}
\noindent for the vectors satisfying $\sum_i{x}_i = \sum_i{y}_i = 1$ and ${x}_i \geq 0$, ${y}_i \geq 0$ for all $1 \leq i \leq \mathcal{D}$. Hence, we apply a small transformation while constructing the  kernel matrix of the SVM in order to approximate the \textit{chi} similarity, instead of using $\text{\normalfont cos}^{-1}\rho_{\chi^2} / \pi$ in Eq. \eqref{eq:main}, which is estimated directly via the Hamming distance.

\subsubsection{Experimental results}
For the multi-label node classification task, Tables \ref{tab:classification-blogcatalog}-\ref{tab:classification-youtube} report the average Micro-$F_1$ and Macro-$F_1$ scores over $10$ runs, where the experiments are performed on different training set sizes. The symbol "-" is used to indicate that the corresponding algorithm is unable to run due to  excessive memory usage ($>$ $128$GB) or because it requires more than one day to complete. The best and second best performing models for each training ratio ($10\%$, $50\%$, and $90\%$) are indicated with bold and underlined text, respectively.
\begin{table}
\centering
\resizebox{.95\columnwidth}{!}{
\begin{tabular}{@{}rcccccc@{}}\toprule
\multicolumn{1}{l}{} & \multicolumn{3}{c}{Micro-$F_1$} & \multicolumn{3}{c}{Macro-$F_1$} \\\midrule
\multicolumn{1}{r}{} & \textbf{10\%} & \textbf{50\%} & \multicolumn{1}{c}{\textbf{90\%}} & \textbf{10\%} & \textbf{50\%} & \textbf{90\%} \\ \cmidrule(lr){1-1}\cmidrule(lr){2-4}\cmidrule(lr){5-7}
\textsc{HOPE} & 0.305 & 0.317 & 0.326 & 0.117 & 0.119 & 0.124 \\
\textsc{Node2Vec} & 0.341 & 0.352 & 0.345 & 0.155 & 0.165 & 0.165 \\
\textsc{NetSMF} & \textbf{0.360} & 0.376 & 0.377 & {\ul 0.189} & 0.200 & 0.199 \\
\textsc{FREDE} & 0.354 & 0.368 & 0.381 & 0.171 & 0.179 & 0.183 \\
\textsc{LouvainNE} & 0.047 & 0.143 & 0.165 & 0.022 & 0.037 & 0.041 \\
\textsc{RandNE} & 0.316 & 0.337 & 0.340 & 0.141 & 0.164 & 0.165 \\
\textsc{NodeSketch} & 0.305 & {\ul 0.381} & {\ul 0.398} & 0.145 & {\ul 0.236} & {\ul 0.263} \\
\textsc{\modelabbrv} & {\ul 0.358} & \textbf{0.408} & \textbf{0.420} & \textbf{0.191} & \textbf{0.267} & \textbf{0.286} \\ \bottomrule
\end{tabular}
}
\caption{Micro-$F_1$ and Macro-$F_1$ classification scores  for varying training set ratios of the \textsl{Blogcatalog} network.}
\label{tab:classification-blogcatalog}
\end{table}
\begin{table}
\centering
\resizebox{.95\columnwidth}{!}{
\begin{tabular}{@{}rcccccc@{}}\toprule
\multicolumn{1}{l}{} & \multicolumn{3}{c}{Micro-$F_1$} & \multicolumn{3}{c}{Macro-$F_1$} \\\midrule
\multicolumn{1}{r}{} & \textbf{10\%} & \textbf{50\%} & \multicolumn{1}{c}{\textbf{90\%}} & \textbf{10\%} & \textbf{50\%} & \textbf{90\%} \\ \cmidrule(lr){1-1}\cmidrule(lr){2-4}\cmidrule(lr){5-7}
\textsc{HOPE} & 0.687 & 0.780 & 0.797 & 0.671 & 0.772 & 0.786 \\
\textsc{Node2Vec} & {\ul 0.764} & 0.813 & 0.831 & 0.749 & 0.802 & 0.818 \\
\textsc{NetSMF} & 0.763 & 0.824 & 0.831 & 0.751 & 0.815 & 0.821 \\
\textsc{FREDE} & \textbf{0.777} & {\ul 0.825} & 0.846 & \textbf{0.766} & 0.817 & 0.833 \\
\textsc{LouvainNE} & 0.686 & 0.711 & 0.721 & 0.648 & 0.675 & 0.683 \\
\textsc{RandNE} & 0.583 & 0.676 & 0.693 & 0.557 & 0.668 & 0.686 \\
\textsc{NodeSketch} & 0.648 & {\ul 0.825} & {\ul 0.872} & 0.632 & {\ul 0.818} & {\ul 0.866} \\
\textsc{\modelabbrv} & 0.750 & \textbf{0.852} & \textbf{0.879} & {\ul 0.736} & \textbf{0.843} & \textbf{0.871} \\ \bottomrule
\end{tabular}
}
\caption{Micro-$F_1$ and Macro-$F_1$ classification scores for varying training set ratios of the \textsl{Cora} network.}
\label{tab:classification-cora}
\end{table}
\begin{table}
\centering
\resizebox{.95\columnwidth}{!}{
\begin{tabular}{rcccccc}\toprule
\multicolumn{1}{l}{} & \multicolumn{3}{c}{Micro-$F_1$} & \multicolumn{3}{c}{Macro-$F_1$} \\\midrule
\multicolumn{1}{r}{} & \textbf{\textbf{10\%}} & \textbf{\textbf{50\%}} & \multicolumn{1}{c}{\textbf{90\%}} & \textbf{\textbf{10\%}} & \textbf{\textbf{50\%}} & \textbf{\textbf{90\%}} \\ \cmidrule(lr){1-1}\cmidrule(lr){2-4}\cmidrule(lr){5-7}
\textsc{HOPE} & 0.620 & 0.632 & 0.631 & 0.525 & 0.536 & 0.536 \\
\textsc{Node2Vec} & 0.621 & 0.632 & 0.631 & 0.510 & 0.535 & 0.531 \\
\textsc{NetSMF} & 0.626 & 0.644 & 0.647 & 0.533 & 0.572 & 0.575 \\
\textsc{FREDE} & 0.648 & 0.661 & 0.661 & 0.567 & 0.586 & 0.588 \\
\textsc{LouvainNE} & 0.494 & 0.496 & 0.499 & 0.354 & 0.356 & 0.359 \\
\textsc{RandNE} & 0.418 & 0.437 & 0.438 & 0.233 & 0.255 & 0.257 \\
\textsc{NodeSketch} & {\ul 0.668} & \textbf{0.847} & \textbf{0.903} & {\ul 0.616} & \textbf{0.831} & \textbf{0.891} \\
\textsc{\modelabbrv} & \textbf{0.704} & {\ul 0.843} & {\ul 0.893} & \textbf{0.660} & {\ul 0.824} & {\ul 0.879} \\ \bottomrule
\end{tabular}
}
\caption{Micro-$F_1$ and Macro-$F_1$ classification scores for  varying training set ratios of the \textsl{DBLP} network.} 
\label{tab:classification-dblp}
\end{table}
\begin{table}
\centering
\resizebox{.95\columnwidth}{!}{
\begin{tabular}{rcccccc}\toprule
\multicolumn{1}{l}{} & \multicolumn{3}{c}{Micro-$F_1$} & \multicolumn{3}{c}{Macro-$F_1$} \\\midrule
\multicolumn{1}{r}{} & \textbf{\textbf{10\%}} & \textbf{\textbf{50\%}} & \multicolumn{1}{c}{\textbf{90\%}} & \textbf{\textbf{10\%}} & \textbf{\textbf{50\%}} & \textbf{\textbf{90\%}} \\ \cmidrule(lr){1-1}\cmidrule(lr){2-4}\cmidrule(lr){5-7}
\textsc{HOPE} & 0.134 & 0.151 & 0.146 & 0.083 & 0.085 & 0.077 \\
\textsc{Node2Vec} & 0.141 & 0.161 & 0.138 & 0.084 & 0.087 & 0.070 \\
\textsc{NetSMF} & 0.150 & 0.170 & 0.163 & 0.096 & 0.102 & 0.095 \\
\textsc{FREDE} & {\ul 0.156} & 0.174 & 0.157 & 0.099 & 0.105 & 0.090 \\
\textsc{LouvainNE} & 0.042 & 0.054 & 0.056 & 0.023 & 0.025 & 0.021 \\
\textsc{RandNE} & 0.145 & 0.161 & 0.145 & 0.087 & 0.091 & 0.083 \\
\textsc{NodeSketch} & 0.152 & {\ul 0.227} & {\ul 0.243} & {\ul 0.102} & {\ul 0.181} & \textbf{0.196} \\
\textsc{\modelabbrv} & \textbf{0.177} & \textbf{0.236} & \textbf{0.246} & \textbf{0.119} & \textbf{0.185} & {\ul 0.191} \\ \bottomrule
\end{tabular}
}
\caption{Micro-$F_1$ and Macro-$F_1$ classification scores for varying training set ratios of the \textsl{PPI} network.}
\label{tab:classification-ppi}
\end{table}
\begin{table}
\centering
\resizebox{.95\columnwidth}{!}{
\begin{tabular}{rcccccc}\toprule
\multicolumn{1}{l}{} & \multicolumn{3}{c}{Micro-$F_1$} & \multicolumn{3}{c}{Macro-$F_1$} \\\midrule
 & \textbf{\textbf{10\%}} & \textbf{\textbf{50\%}} & \textbf{\textbf{90\%}} & \textbf{\textbf{10\%}} & \textbf{\textbf{50\%}} & \textbf{\textbf{90\%}} \\ \cmidrule(lr){1-1}\cmidrule(lr){2-4}\cmidrule(lr){5-7}
\textsc{HOPE} & 0.342 & 0.341 & 0.343 & 0.198 & 0.201 & 0.201 \\
\textsc{Node2Vec} & - & - & - & - & - & - \\
\textsc{NetSMF} & 0.392 & 0.379 & 0.376 & 0.273 & 0.256 & 0.247 \\
\textsc{FREDE} & - & - & - & - & - & - \\
\textsc{LouvainNE} & 0.248 & 0.251 & 0.256 & 0.064 & 0.063 & 0.072 \\
\textsc{RandNE} & 0.335 & 0.341 & 0.339 & 0.205 & 0.220 & 0.215 \\
\textsc{NodeSketch} & {\ul 0.439} & \textbf{0.467} & \textbf{0.476} & {\ul 0.365} & \textbf{0.412} & \textbf{0.426} \\
\textsc{\modelabbrv} & \textbf{0.455} & {\ul 0.465} & {\ul 0.471} & \textbf{0.387} & {\ul 0.410} & {\ul 0.414}\\\bottomrule
\end{tabular}
}
\caption{Micro-$F_1$ and Macro-$F_1$ classification scores for varying training set ratios of  the \textsl{Youtube} network.}
\label{tab:classification-youtube}
\end{table}

As we observe, \textsc{\modelabbrv} consistently outperforms the baselines for higher training ratios on the \textsl{Blogcatalog} and \textsl{Cora} networks, while the  obtained Macro-$F_1$ score is very close  to the performance of \textsc{NetSMF} for $10\%$ training ratio on \textsl{Blogcatalog}. In the case of the \textsl{Cora} network which corresponds to the smallest one used in our study, \textsc{FREDE} shows better performance for small training ratio of $10\%$. For the \textsl{Youtube} and \textsl{DBLP} networks, the proposed \textsc{\modelabbrv} model along with \textsl{NodeSketch} perform equally well. This is quite surprising, since both these methods that correspond to data-independent hashing techniques offer a clear performance gain over traditional models, such as \textsc{Node2Vec} and HOPE. Lastly, for the \textsl{PPI} dataset, \textsc{\modelabbrv} obtains consistently the highest scores for Micro-$F_1$, while its main competitor \textsc{NodeSketch} has close performance for the Macro-$F_1$ score.

\subsection{Link Prediction}
The second downstream task used  to assess the quality of node embeddings is the one of link prediction.

\subsubsection{Experimental set-up}
Half of the edges of a given network are removed by still keeping the residual network connected. Node embeddings are learned on the rest of the graph. The removed edges are considered as positive samples for the testing set, while the same number of node pairs which does not exist in the initial network is separately sampled for training and testing sets in order to form the negative samples. As it has been described in Section \ref{subsec:classification}, we build the features corresponding to the node pair samples using the similarities between embedding vectors; the similarity measure is chosen depending on the algorithm that we use to extract the representations. Since \textsl{Youtube} is relatively larger than the rest of the networks, we work on $7\%$ of its initial size. We predict edges by constructing the similarity list of edges, and we provide the \textit{Area Under Curve} (AUC) scores in Table \ref{tab:link_pred}.
\begin{table}[]
\centering
\resizebox{\columnwidth}{!}{%
\begin{tabular}{rccccc}
\toprule
\textbf{} & \multicolumn{1}{l}{Blogcatalog} & \multicolumn{1}{l}{Cora} & \multicolumn{1}{l}{DBLP} & \multicolumn{1}{l}{PPI} & \multicolumn{1}{l}{Youtube} \\ \midrule
\textsc{HOPE} & 0.517 & 0.665 & 0.769 & 0.524 & 0.514 \\
\textsc{Node2Vec} & 0.595 & {\ul 0.748} & 0.843 & {\ul 0.616} & 0.533 \\
\textsc{NetSMF} & 0.691 & 0.709 & 0.835 & 0.534 & \textbf{0.542} \\
\textsc{FREDE} & {\ul 0.709} & \textbf{0.760} & \textbf{0.858} & 0.451 & 0.460 \\
\textsc{LouvainNE} & 0.565 & 0.684 & 0.789 & 0.570 & 0.528 \\
\textsc{RandNE} & 0.608 & 0.508 & 0.517 & 0.505 & 0.502 \\
\textsc{NodeSketch} & 0.703 & 0.590 & 0.714 & 0.514 & 0.510 \\
\textsc{\modelabbrv} & \textbf{0.822} & 0.737 & {\ul 0.856} & \textbf{0.654} & {\ul 0.537} \\ \bottomrule
\end{tabular}%
}
\caption{Area Under Curve scores for link prediction.}
\label{tab:link_pred}
\end{table}

\subsubsection{Experimental results}
For the link prediction task, \textsc{\modelabbrv} acquires the highest AUC scores on three datasets, while it is also the second-best performing model for the remaining two. In the case of the \textsl{Youtube} dataset, all  baselines demonstrate comparable results. Although \textsc{Node2Vec} shows good performance across most datasets in the link prediction task, it does not perform well on the \textsl{Blogcatalog} network, mainly because of its high density. On the other hand, \textsc{\modelabbrv} reaches the highest score on this dataset, with a clear difference to its main competitor, \textsc{NodeSketch}.

\subsection{Parameter Sensitivity}\label{subsection:param_sensitivity}
We concentrate on the influence of three parameters, namely walk length $L$, importance factor $\alpha$ and dimension size $\mathcal{D}$, examining their impacts on the \textsl{Cora} network.

\subsubsection{Effect of walk length}
\begin{figure}[h!]
	\centering
	\includegraphics[width=\linewidth]{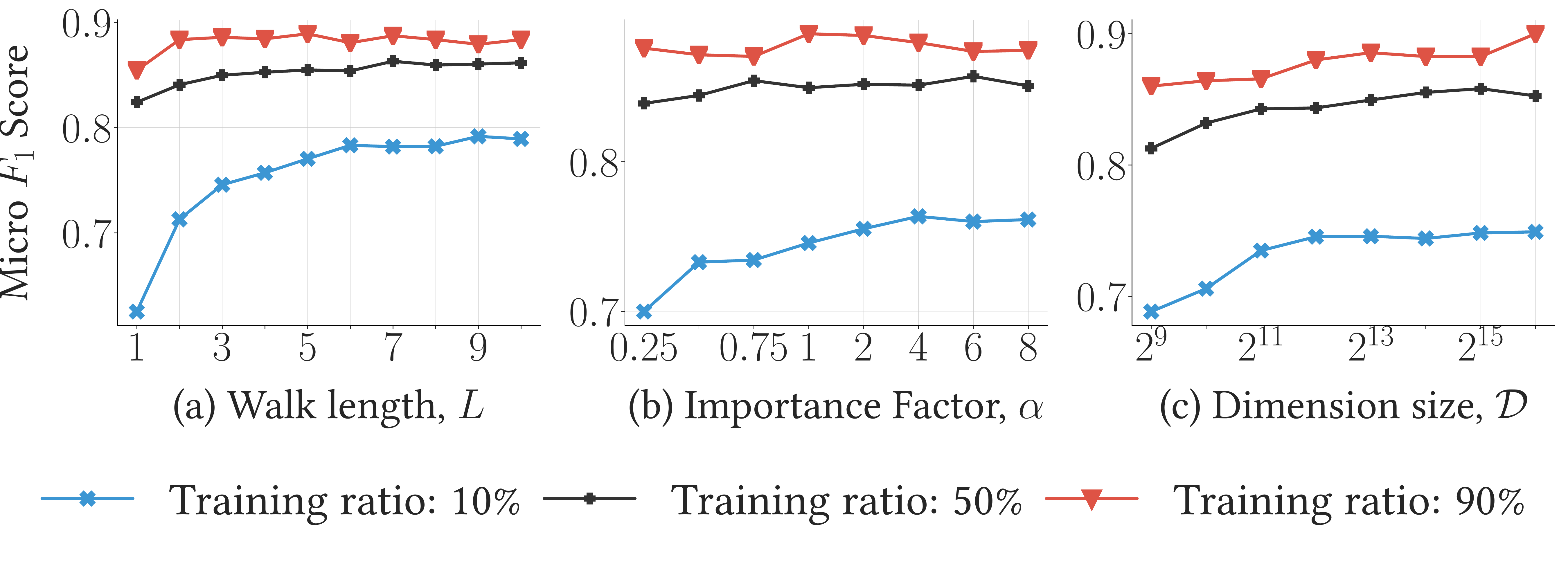}
	\caption{Influence of various parameters in terms of Micro-$F_1$ score on the \textsl{Cora} network for varying training set ratios.}\label{fig:example}
\end{figure}

In order to examine the influence of the walk length on the performance, we perform experiments for varying lengths by fixing the importance factor $\alpha$ to $1.0$. Figure \ref{fig:example}a depicts the Micro-$F_1$ scores  for different training ratios.  We observe a significant increase in performance when the walk length increases, particularly for small training ratios and walk lengths. Although it shows a wavy behavior for the largest training ratio, there is a logarithmic improvement depending on the walk length. \textsc{\modelabbrv} better captures the structural properties of the network in longer walks, thus the low performance  observed on small training ratios can be compensated with longer walks.

\subsubsection{Effect of importance factor}
The importance factor controls the impact of walks of different lengths: the importance of the higher levels is increasing for $\alpha > 1$, while it can be diminished choosing $\alpha < 1$. Figure \ref{fig:example}b  depicts the performance of \textsc{\modelabbrv} on the \textsl{Cora} network,  fixing the walk length value to $3$. Although we do not observe a steady behavior for the large training set, higher values of $\alpha$, especially around $4$, positively contribute to the performance;  values smaller than $1$ have negative impact on the performance.

\subsubsection{Effect of dimension size}
The dimension size is a crucial parameter affecting the performance of the algorithm, since a better approximation to the $chi$ similarity measure can be obtained for larger dimension sizes, following \textit{Hoeffding's} inequality \cite{hoeffding}. Therefore, we perform experiments for varying dimension sizes, by fixing the walk length to $5$. Figure \ref{fig:example}c depicts the Micro-$F_1$ scores of the classification experiment for different dimension sizes ranging from $2^9$ to $2^{17}$. Although we have fluctuating scores on the large training set due to the randomized behavior of the approach, the impact of the dimension size can be observed clearly on the small training set size. On the other hand, we observe an almost stable behavior for the training ratio of $50\%$, encouraging the use of small embedding sizes towards reducing storage requirements. 

\subsection{Time Comparison}
We have recorded the elapsed real (wall clock) time of all methods, and the results are provided in Table \ref{tab:runningtime}. The \textit{Random} network indicates the  $\mathcal{G}_{n,p}$ Erd\"os-Renyi random graph model, using $n=10^5$ and $p=10^{-4}$. All the experiments have been conducted on the server whose specifications given in Section \ref{sec.performance}. We use $32$ threads for each algorithm, when it is applicable. We have utilized the suggested default parameters for the baselines, and the settings described for the classification task are employed for \textsc{\modelabbrv}.
\begin{table}[h!]
\centering
\resizebox{\columnwidth}{!}{%
\begin{tabular}{r|cccccc|c}
 & \textbf{ \rotatebox{75}{\textsl{Blogcatalog}}} & \textbf{\rotatebox{75}{\textsl{Cora}}} & \textbf{\rotatebox{75}{\textsl{DBLP}}} & \textbf{\rotatebox{75}{\textsl{PPI}}} & \textbf{\rotatebox{75}{\textsl{Youtube}}} & \textbf{\rotatebox{75}{\textsl{Random}}} & \textbf{\rotatebox{75}{Speedup}}  \\\cmidrule(lr){1-1}\cmidrule(lr){2-7}\cmidrule(lr){8-8}
\multicolumn{1}{r|}{\textsc{HOPE}} & 97.81 & 27.32 & 198.59 & 32.65 & 8470.33 & 1048.52 & 8.85x \\
\multicolumn{1}{r|}{\textsc{Node2Vec}} & 1400.44 & 18.32 & 161.24 & 72.16 & - & 716.07 & 2.55x \\
\multicolumn{1}{r|}{\textsc{NetSMF}} & 7.78 & 1.32 & 10.91 & 1.90 & 1624.94 & 236.30 & 1.69x \\
\multicolumn{1}{r|}{\textsc{FREDE}} & 1179.79 & 20.46 & 2612.84 & 140.43 & - & 22386.98 & 28.33x \\
\multicolumn{1}{r|}{\textsc{LouvainNE}} & 0.34 & 0.06 & 0.24 & 0.11 & 6.86 & 1.25 & 0.01x \\
\multicolumn{1}{r|}{\textsc{RandNE}} & 25.52 & 3.15 & 11.55 & 5.40 & 449.15 & 73.11 & 0.51x \\
\multicolumn{1}{r|}{\textsc{NodeSketch}} & 64.21 & 13.42 & 19.10 & 14.40 & 1563.00 & 101.16 & 1.59x \\
\multicolumn{1}{r|}{\textsc{\modelabbrv}} & 17.40 & 0.74 & 9.26 & 2.53 & 1047.53 & 38.11 & 1.00x\\\bottomrule
\end{tabular}%
}
\caption{Running time (in seconds) and average speedup.}
\label{tab:runningtime}
\end{table}

As we observe, \textsc{\modelabbrv} runs faster than \textsc{HOPE}, \textsc{Node2Vec} as well as \textsc{FREDE}. This is happening because \textsc{HOPE} requires an expensive matrix factorization, while \textsc{Node2Vec} needs to simulate random walks to obtain their exact realizations. 
Although \textsc{FREDE} is a sketching-based approach, we have observed that the computation of the PPR matrix requires considerable time.
Furthermore, although the remaining baseline methods run faster compared to \textsc{\modelabbrv}, as we have already presented, the proposed model generally outperforms them both in the classification and link prediction tasks. These experiments further support the intuition about designing \textsc{\modelabbrv} as an expressive model that balances accuracy and running time. 




%% file: 5-discussion.tex
\section{Discussion for Dynamic Networks}\label{sec:discussion}
Most real-world networks undergo structural changes and evolve over time with the addition and removal of links and nodes \cite{JMLR:v21:19-447}. Therefore, designing models properly adapting to dynamic networks is an important point to investigate. As we discuss here,  the proposed method allows for efficient updates of the embeddings, without requiring any costly learning procedures. 
More precisely, the key point in the dynamic case, is that the learned embedding vectors should be efficiently updated instead of being recalculated from scratch. If an edge is added or removed for a pair of nodes $(u,v)\in V \times V$, the terms $\mathcal{R}^{(l)}_{(w, :)}$ in Eq. \eqref{eq:update_rule} for node $w\in V$ are affected, for all $l > k :=\min\{dist(w,u), dist(w,v)\}$---thus, it suffices to update  only these affected terms. The transition probabilities for nodes $u$ and $v$ also change even though the remaining nodes are not affected, so all the terms $P_{(v,:)}$ must be divided by $\sum_{w\in\mathcal{N}(v)}P_{(v,w)}$ in order to normalize the transition probabilities and similarly the same procedure must also be applied to node $u$ after each edge insertion and deletion operation.

%% file: 6-conclusion.tex
\section{Conclusion and Future Work}
\label{sec.conclusions}
We have introduced \textsc{\modelabbrv}, an efficient binary node embedding model. Its components have properly been designed to improve scalability without sacrificing effectiveness on downstream tasks. \textsc{\modelabbrv} exploits random walk diffusion probabilities via stable random projection hashing, towards  efficiently computing representations in the Hamming space that approximate the \textit{chi} similarity. The experimental results have demonstrated that \textsc{\modelabbrv} outperformed in accuracy recent highly-scalable models, being able to run within the reasonable time duration, while at the same time it shows comparable or even better accuracy with respect to widely used baseline methods in multi-label node classification and link prediction.
In future work, we plan to further study the properties of the model for attributed and dynamic networks and also study the performance of parallel/distributed alternatives.